\documentclass[letterpaper]{article} 
\usepackage{aaai2026}  
\usepackage{times}  
\usepackage{helvet}  
\usepackage{courier}  
\usepackage[hyphens]{url}  
\usepackage{graphicx} 
\usepackage{natbib}  
\usepackage{caption} 
\usepackage{algorithm}
\usepackage{algorithmic}
\usepackage{amsmath}
\usepackage{amssymb} 
\usepackage{xcolor}
\usepackage[table]{xcolor}
\usepackage{multirow}
\usepackage{makecell}
\usepackage{booktabs}

\usepackage{newfloat}
\usepackage{listings}
\DeclareCaptionStyle{ruled}{labelfont=normalfont,labelsep=colon,strut=off} 
\floatstyle{ruled}
\newfloat{listing}{tb}{lst}{}
\floatname{listing}{Listing}
\title{Anatomical Region-Guided Contrastive Decoding: A Plug-and-Play Strategy for Mitigating Hallucinations in Medical VLMs}
\author{
    Xiao Liang,\textsuperscript{\rm 1}
    Chenxi Liu,\textsuperscript{\rm 1}
    Zhi Ma,\thanks{Corresponding authors.}\textsuperscript{\rm 1}
    Di Wang,\footnotemark[1]\textsuperscript{\rm 1}
    Bin Jing,\textsuperscript{\rm 2}
    Quan Wang,\textsuperscript{\rm 1}
    Yuanyuan Shi\textsuperscript{\rm 3}
}
\affiliations{
    \textsuperscript{\rm 1}School of Computer Science and Technology, Xidian University, China\\
    \textsuperscript{\rm 2}School of Biomedical Engineering, Capital Medical University, China\\
    \textsuperscript{\rm 3}Department of Ophthalmology, the Ninth Medical Center of the Chinese PLA General Hospital, China\\
    ecoxial2012@outlook.com
}

\usepackage{bibentry}

\begin{document}

\maketitle

\begin{abstract}
Medical Vision-Language Models (MedVLMs) show immense promise in clinical applicability. However, their reliability is hindered by hallucinations, where models often fail to derive answers from visual evidence, instead relying on learned textual priors. Existing mitigation strategies for MedVLMs have distinct limitations: training-based methods rely on costly expert annotations, limiting scalability, while training-free interventions like contrastive decoding, though data-efficient, apply a global, untargeted correction whose effects in complex real-world clinical settings can be unreliable. To address these challenges, we introduce \textbf{A}natomical \textbf{R}egion-Guided \textbf{C}ontrastive \textbf{D}ecoding \textbf{(ARCD)}, a plug-and-play strategy that mitigates hallucinations by providing targeted, region-specific guidance. Our module leverages an anatomical mask to direct a three-tiered contrastive decoding process. By dynamically re-weighting at the token, attention, and logits levels, it verifiably steers the model's focus onto specified regions, reinforcing anatomical understanding and suppressing factually incorrect outputs. Extensive experiments across diverse datasets, including chest X-ray, CT, brain MRI, and ocular ultrasound, demonstrate our method's effectiveness in improving regional understanding, reducing hallucinations, and enhancing overall diagnostic accuracy.
\end{abstract}

\begin{links}
    \link{Code}{https://github.com/ecoxial2007/ARegionCD}
\end{links}

\section{Introduction}
\label{sec1:intro}
\begin{figure}[t]
  \centering
  \includegraphics[width=1\columnwidth]{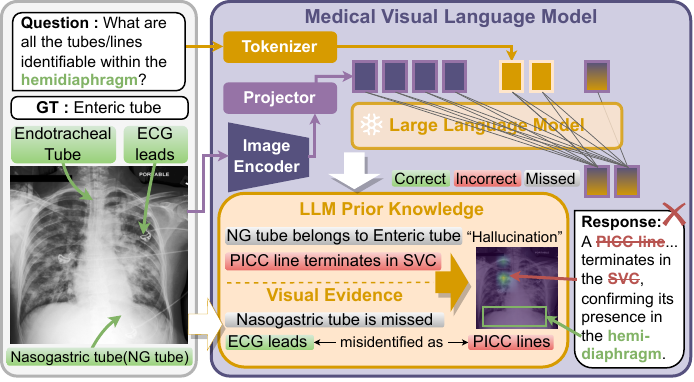}
  \caption{An example of hallucination driven by a statistical bias. The model misidentifies visually apparent \textit{ECG leads} as a \textit{PICC line} because the latter is far more common in training corpora reports. This flawed prior-visual association leads to a factually incorrect response and demonstrates a critical failure in visual grounding.}
  \label{fig:existing_problems}
\end{figure}

In recent years, Medical Vision-Language Models (MedVLMs) have demonstrated impressive capabilities across diverse modalities such as chest radiography \cite{RadFM}, pathology \cite{Seyfioglu2023QuiltLLaVAVI}, and dermatology \cite{Zeng2025MMSkinED}, showing great promise in advancing medical intelligence for tasks like automated reporting and visual question answering. However, the clinical applicability of these models is critically undermined by their propensity for hallucination \cite{Bai2024HallucinationOM}—generating plausible yet factually incorrect statements that contradict the visual evidence. This unreliability stems from a fundamental lack of visual grounding, where models often prioritize statistical biases and unimodal priors over actual image content \cite{vcdLeng2023MitigatingOH}. For instance, a model may misidentify \textit{ECG leads} as a \textit{PICC line}, as shown in Figure \ref{fig:existing_problems}. This confusion often arises because textual reports in its training data frequently describe \textit{PICC lines} but rarely mention the visually similar \textit{ECG leads}, creating a strong statistical prior that overrides the visual evidence. The opaque nature of this failure mechanism raises a critical question for clinical trust: \textit{How can we ensure a VLM's attention is verifiably focused on the specific, diagnostically relevant regions of an image, mirroring the analytical process of a physician?}

To mitigate hallucinations and enhance the visual understanding of VLMs, researchers have pursued two main avenues. The first involves training-based approaches. For instance, methods like MMedPO \cite{zhu2024mmedpo} leverage preference optimization to compel the model to ground its responses in visual evidence rather than relying on textual priors. The second avenue is training-free inference-time intervention. Techniques such as visual contrastive decoding \cite{vcdLeng2023MitigatingOH} work by contrasting outputs from original and distorted images to penalize the model's over-reliance on learned data biases. While these approaches are valuable, they present a difficult trade-off. Training-based methods are constrained by the prohibitively high cost of annotating medical data, which limits their broad applicability. On the other hand, training-free methods, though data-efficient, apply a global, untargeted correction whose effects on clinically complex medical data can be unreliable.

To address these challenges, we propose \textbf{A}natomical \textbf{R}egion-Guided \textbf{C}ontrastive \textbf{D}ecoding \textbf{(ARCD)}, a plug-and-play decoding strategy that provides fine-grained, multi-level guidance on the model's visual understanding, enhancing its reliability on clinically complex data without requiring additional training. We first build upon a strong MedVLM baseline established by fine-tuning Phi-3.5V on high-quality medical visual instruction data. The proposed ARCD strategy then performs \textbf{Dynamic Attention Mask Generation} to convert the segmentation annotation of a given anatomical region into a token-level attentional mask, which in turn directs our \textbf{Mask-Guided Conditional Token Weighting} module. This module steers the decoding process using a three-tiered strategy: it applies dynamic, contrastive re-weighting between guided and unguided generation branches sequentially at the token, attention, and logits levels, which ultimately reinforces anatomical region-specific understanding at inference time, thereby mitigating hallucinations. Our primary contributions are:

\begin{itemize}
    \item We propose \textbf{Anatomical Region-Guided Contrastive Decoding}, a novel decoding strategy that improves visual grounding by guiding language generation with a three-tiered, contrastive mechanism based on specified anatomical regions.
    \item The proposed \textbf{ARCD} method is entirely training-free and plug-and-play, allowing for seamless integration with various VLMs and segmentation models without requiring any parameter updates.
    \item We conduct extensive experiments across four diverse medical modalities (chest X-ray, CT, brain MRI, and ocular B-ultrasound) to validate the effectiveness and generalization of our method.
\end{itemize}

\section{Related work}
\subsection{Medical Visual Language Models}

Recent advancements in Medical Vision-Language Models (Med-VLMs) have primarily focused on adapting general-domain models to the specialized needs of healthcare. Foundational works established this paradigm, with LLaVA-Med using instruction-tuning to align biomedical visual features with language embeddings for cost-effective adaptation \cite{li2023llava}, while HuatuoVision leveraged massive, high-quality VQA datasets to further enhance performance \cite{huatuogptv}. Following this, a trend towards domain specialization emerged. This includes models purpose-built for specific modalities, such as RadFM \cite{RadFM} for chest radiography with 2D/3D capabilities, Quilt-LLaVA \cite{Seyfioglu2023QuiltLLaVAVI} for pathology, which creates spatially-grounded data by tracking narrator cursors in educational videos, and MMSkin \cite{Zeng2025MMSkinED} for dermatology, which uses high-fidelity image-text pairs from professional textbooks. To enhance reliability and mitigate hallucinations, advanced techniques have emerged. These range from using preference optimization for clinical accuracy \cite{zhu2024mmedpo} to integrating weak region-of-interest annotations for better localization \cite{chen2024r}. Concurrently, other methods like rationale-based explanations \cite{gai2024medthink}, latent prompts \cite{gu2024lapa}, feature fusion \cite{ha2024fusion}, and self-training pipelines \cite{sun2024self} have collectively improved model interpretability and robustness. However, despite progress in textual accuracy, a key challenge remains in fine-grained visual grounding, where models often fail to explicitly link generated statements to specific, diagnostically relevant image regions \cite{chen2024r, wu2024hallucination}.

\subsection{Hallucination-Mitigated VLMs}
Object hallucination in general Vision-Language Models (VLMs), where generated text misaligns with visual facts, is a well-studied issue \cite{liu2024survey, lan2024survey}. Its causes include knowledge scarcity from limited data \cite{chen2024low}, an over-reliance on statistical and language priors \cite{favero2024multi, vcdLeng2023MitigatingOH}, and perceptual failures from poor image quality \cite{liu2024rebalancing}. Mitigation strategies are either training-based or training-free. Training-based methods aim for intrinsic correction by enriching data (e.g., ShareGPT4V \cite{chen2023sharegpt4v}), using preference optimization like DPO to favor factual responses \cite{zhao2023hallucination, yang2025mitigating}, fusing visual reasoning \cite{park2025second}, or applying adversarial training \cite{chen2025perturbollava}. Training-free methods intervene at inference, using techniques like Visual and Instructional Contrastive Decoding (VCD \cite{vcdLeng2023MitigatingOH}, ICD \cite{wang2024mitigating}), retrieval-augmented generation \cite{feng2024hyper}, or post-processing correction frameworks such as Woodpecker \cite{yin2024woodpecker}. Despite their effectiveness in general domain, applying these methods to the medical domain presents unique challenges \cite{yan2025medhalltune}. Training-based approaches are often hindered by the prohibitive cost and expertise required for medical data annotation. Meanwhile, training-free methods may lack the high degree of interpretability and the rigorous, verifiable accuracy essential for high-stakes clinical applications.

\begin{figure*}[t]
  \centering
  \includegraphics[width=\textwidth]{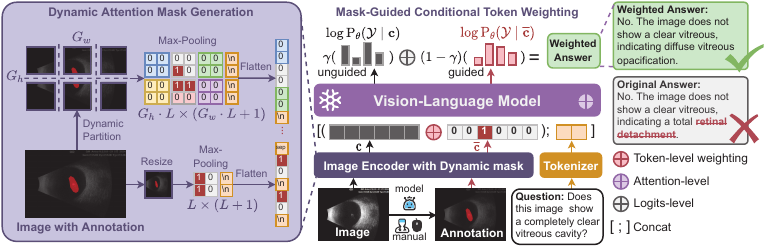} 

\caption{Overview of our proposed Anatomical Region-Guided Contrastive Decoding strategy. \textbf{Left:} \textit{Dynamic Attention Mask Generation} module converts a specified anatomical region (e.g., a segmentation annotation) into a multi-scale token-level mask. \textbf{Right:} \textit{Mask-Guided Conditional Token Weighting} module then uses this mask to steer the decoding process via a strategy that applies contrastive re-weighting at the token level, attention level, and logits level, ensuring the generated answer is grounded in the specified visual region.}

  \label{fig2:overview}
\end{figure*}

\section{Methodology}

Our proposed Anatomical Region-Guided Contrastive Decoding is a plug-and-play strategy that builds upon a strong medically-adapted VLM baseline (Phi3.5V-Med). This module first generates a token-level mask from a spatial annotation that delineates a specific anatomical region (\textit{Dynamic Attention Mask Generation}) and subsequently guides the contrastive decoding process via a three-tiered strategy (\textit{Mask-Guided Conditional Token Weighting}). An overview of our method is illustrated in Figure~\ref{fig2:overview}.

\subsection{Task Formulation and Baseline}

A Vision-Language Model (VLM) aims to generate a coherent and contextually relevant text sequence $\mathcal{Y} = \{y_1, y_2, ..., y_T\}$ in response to a given image $\mathcal{V}$ and a textual prompt or question $\mathcal{Q}$. The generation process is typically auto-regressive, where the objective is to maximize the guided log-probability of the target sequence. This is formally expressed as:
\begin{equation}
\log \mathrm{P}_{\theta}(\mathcal{Y}|\mathcal{V}, \mathcal{Q}) = \sum_{t=1}^{T} \log \mathrm{P}_{\theta}(y_t | \mathcal{V}, \mathcal{Q}, y_{<t}),
\end{equation}
where $y_t$ is the token at timestep $t$, and $y_{<t}$ represents all previously generated tokens. While powerful, lightweight models like Phi-3.5 Vision \cite{Abdin2024Phi3TR} possess strong general visual understanding capabilities, they lack the specialized domain knowledge required for medical applications. Consequently, they struggle to answer medical questions accurately and often fail to identify specific pathologies, even when they are visually prominent. To create a robust baseline, we must inject medical expertise into the model. We perform parameter-efficient fine-tuning on Phi-3.5 Vision using Low-Rank Adaptation (LoRA) \cite{Hu2021LoRALA} with the large-scale PubMedVision \cite{huatuogptv} dataset. This process enriches the model with essential medical knowledge and aligns its outputs with clinical context. The resulting model, which we term Phi3.5V-Med, serves as our baseline for all subsequent experiments.

\subsection{Dynamic Attention Mask Generation}

Despite infusing medical knowledge into VLMs through visual instruction fine-tuning, the data-driven nature of Supervised Fine-Tuning (SFT) makes it challenging to guarantee that the VLM's responses are genuinely grounded in the provided visual evidence, as highlighted in Section 1. To address this, we propose our Anatomical Region-Guided Contrastive Decoding strategy, aiming to explicitly direct the model's focus towards clinically relevant anatomical regions. Specifically, an expert or a pre-trained segmentation model (e.g., PSPNet or MedSAM) first annotates or generates a segmentation mask $\mathcal{S}$ of the same spatial dimensions as the input image $\mathcal{V} \in \mathbb{R}^{H \times W \times C}$. This mask $\mathcal{S} \in \{0, 1\}^{H \times W}$ defines the anatomical region of interest, with pixel values of 1 indicating the relevant region. Subsequently, this region-specific information must be effectively injected into the MedVLM's attention mechanism by transforming it into a token-level mask that aligns with the image token embeddings produced by the VLM's vision encoder.

To achieve this, we introduce \textbf{Dynamic Attention Mask Generation}, a module that creates attention masks structurally aligned with the VLM's visual tokens. This approach allows for handling dynamic input resolutions and capturing fine-grained details by generating two distinct masks: a global mask $\mathcal{M}_g$ and a local mask $\mathcal{M}_l$. Specifically, the ViT-L/14 vision encoder from Phi-3.5V processes a $336 \times 336$ input image by dividing it into a $24 \times 24$ grid of $14 \times 14$ pixel patches. This grid is subsequently reshaped into a $12 \times 12$ feature grid, and we denote this dimension as $L=12$. The global mask $\mathcal{M}_g$ is thus derived by downsampling the high-resolution mask $\mathcal{S}$ to a feature-level grid of size $L \times L$. A newline separator $m_{sep}$ (mask value 0) is then appended to each of its $L$ rows, resulting in a flattened one-dimensional mask $\mathcal{M}_g$ of length $L \times (L+1)$. Similarly, the local mask $\mathcal{M}_l$ is generated by downsampling $\mathcal{S}$ to a larger composite grid of size $(G_h \cdot L) \times (G_w \cdot L + 1)$. Here, the user-defined grid dimension $G=(G_h, G_w)$ determines the granularity of the local analysis. A larger $G$ partitions the image into a finer-grained $G_h \times G_w$ grid of local views, allowing the model to focus on smaller details. Subsequently, a newline separator is appended to each of the $G_h \cdot L$ rows of this composite grid before it is flattened into the final sequence $\mathcal{M}_l$. Finally, the complete mask $\mathcal{M}$ is assembled by concatenating the local mask, a single separator token $m_{sep}=(0)$, and the global mask: $\mathcal{M} = [\mathcal{M}_l ; m_{sep} ; \mathcal{M}_g] \in \mathbb{R}^{N}$. This produces a composite and structured attention mask with a total length $N$:
\begin{equation}
    N = G_h \cdot L \cdot (G_w \cdot L + 1) + 1 + L \cdot (L+1).
\end{equation} This mask $\mathcal{M}$ then takes effect during the model's self-attention computations, thereby prioritizing information from the specified anatomical regions during response generation.

\subsection{Mask-Guided Conditional Token Weighting}
While VCD \cite{vcdLeng2023MitigatingOH} mitigates language priors by merging an ``\textit{original}'' and a noise-augmented ``\textit{distorted}'' branch at the logits level, its original branch still contains redundant visual information that can lead to hallucinations. Additionally, this post-hoc intervention at the logits level fails to correct the decoder's erroneous focus on irrelevant visual regions during the attention computation stage. 

To address these limitations, we extend VCD with a multi-level guidance mechanism, fusing a ``\textit{original (unguided)}'' branch $\overline{\mathbf{c}}$ and a ``\textit{guided}'' branch $\mathbf{c}$ via weighted fusion at the token, attention, and logits levels. The conditional guidance for the branch $\mathbf{c}$ is provided by our previously generated visual token mask $\mathcal{M}=\{m_1, m_2, \dots\}$, where the binary indicator $m_i \in \{0, 1\}$ is set to 1 for tokens corresponding to regions requiring enhanced focus. First, the input image $\mathcal{V}$ and question $\mathcal{Q}$ are processed by a tokenizer and encoder $f(\cdot)$ to produce token embeddings. For clarity, we denote the input tokens collectively as $\mathbf{x}$ and the resulting embeddings as $\mathbf{c} = f(\mathbf{x})$. 
At the token-level, a small weight $\alpha$ (e.g., 0.01) is used to suppress the embeddings of the guided tokens within the unguided branch $\overline{\mathbf{c}}$:
\begin{equation}
\label{eq:token_weighting}
\overline{c}_i =
\begin{cases}
  \alpha \cdot c_i & \text{if } m_i = 1 \\
  c_i & \text{if } m_i = 0
\end{cases}
\end{equation}Subsequently, at the attention-level, a weight $\beta$ (e.g., 3, where a larger value implies stronger guidance) is applied to amplify the attention probability $p_i$ corresponding to the visual regions of interest:

\begin{equation}
\label{eq:attention_weighting}
p_i = \frac{\beta^{m_i} \cdot \exp(e_i)}{\sum_{j=1}^{N} \beta^{m_j} \cdot \exp(e_j)}.
\end{equation}
Here, $e_i$ represents the $i$-th element of the pre-softmax query-key attention score matrix within a single attention head, and $N$ is the number of tokens in the input sequence. This step is designed to counteract the tendency for textual information to dominate, ensuring that visual cues are not diminished before being processed by the Large Language Model. Finally, at the logits-level, the parameter $\gamma$ (e.g., 1.5, where a larger value implies stronger guidance) controls the intensity of the guidance towards the guided branch:
\begin{equation}
\label{eq:logits_weighting}
\mathbb{P}_\mathcal{Y} = (1-\gamma) \log \mathrm{P}_{\theta}(\mathcal{Y}|\overline{\mathbf{c}}) + \gamma \log \mathrm{P}_{\theta}(\mathcal{Y}|\mathbf{c}).
\end{equation}Here, $\mathbb{P}_\mathcal{Y}$ represents the final, guided log-probability distribution for the next token, from which the guided response $\mathcal{Y}$ is generated auto-regressively at each decoding step. Through this three-tiered weighted fusion, we achieve a comprehensive and controllable mechanism for anatomical region-guided generation. Beyond steering focus towards visual regions, this framework can also be adapted for textual alignment by modifying the mask $\mathcal{M}$, ensuring that generated content more closely adheres to specific topics \cite{Zhang2023PromptHI}. The optimal hyperparameter settings are detailed in the implementation section.

\section{Experiments}
\subsection{Datasets}
For our experimental setup, we leveraged \textbf{PubMedVision} \cite{huatuogptv}, a large dataset containing 1.3 million MedVQA instances for visual instruction tuning. We then used this to fine-tune the Phi-3.5V model, which served as our zero-shot baseline for subsequent evaluations.  Our downstream tasks encompassed three distinct types of MedVQA challenges. These included \textbf{MIMIC-Ext-VQA} \cite{Bae2023EHRXQAAM} for chest X-rays and \textbf{SLAKE} \cite{SlakeAS_Liu2021_23}, a comprehensive radiology dataset covering abdominal CT, chest X-rays, and brain MRI. Additionally, we incorporated a specialized Ocular B-ultrasound dataset, named \textbf{OBScan}, specifically constructed by us. OBScan is composed of ocular B-scan ultrasound images from patients with various eye diseases, each paired with an expert-authored diagnostic report. These reports are structured into three sections: a description of the left eye, a description of the right eye, and a final impression. To provide granular grounding for the diagnoses, experts manually segmented key anatomical and pathological regions in each image, including the retina, lens, vitreous opacity, and vitreous detachment. Following the methodology of MIMIC-Ext-VQA, we then formulated a diverse set of QA pairs featuring three question formats (\textit{choose}, \textit{verify}, and \textit{query}) and two answer types (\textit{closed} and \textit{open-ended}) for our experiments. The fine-tuning baseline underwent separate fine-tuning processes on each of these three diverse downstream datasets. Further details for all datasets are provided in the Appendix.

\begin{table*}[t]
  \centering
  \resizebox{1\textwidth}{!}{
    \begin{tabular}{p{2cm}ccccccccc}
    \toprule
    \multicolumn{1}{l}{Method} & \multicolumn{3}{c}{\textbf{MIMIC-Ext-VQA*}} & \multicolumn{3}{c}{\textbf{SLAKE}} & \multicolumn{3}{c}{\textbf{OBScan}} \\
\cmidrule{2-10}          & Open  & Closed & Overall & Open  & Closed & Overall & Open  & Closed & Overall \\
    \midrule
    \multicolumn{10}{l}{\textit{General Visual-Language Model}} \\
    \multicolumn{1}{l}{LLaVA-1.6-7B} & 12.70 & 57.07 & 46.06 & 32.28 & 53.54 & 42.91 & 47.78 & 62.80 & 57.48 \\
    \multicolumn{1}{l}{Qwen-VL-7B} & 15.87 & 56.54 & 46.46 & 39.37 & 61.42 & 50.39 & 48.89 & 71.95 & 63.78 \\
    \multicolumn{1}{l}{Phi3.5V-4.2B} & 17.46 & 56.02 & 46.46 & 33.07 & 53.54 & 43.31 & 40.00 & 56.10 & 50.39 \\
    \multicolumn{1}{l}{GPT-4o} & 28.57 & 74.87 & 63.39 & 38.58 & 74.80 & 56.69 & 72.22 & 70.73 & 71.26 \\
    \midrule
    \multicolumn{10}{l}{\textit{Medical Visual-Language Model (Zero-Shot)}} \\
    \multicolumn{1}{l}{LLaVA-Med-7B} & 14.29 & 55.50 & 45.28 & 37.80 & 55.12 & 46.46 & 52.22 & 60.37 & 57.48 \\
    \multicolumn{1}{l}{HuatuoV-34B} & 17.46 & 65.45 & 53.54 & 41.73 & 74.02 & 57.87 & 60.00 & 67.68 & 64.96 \\
    \multicolumn{1}{l}{Phi3.5V-Med} & 14.29 & 58.12 & 47.24 & 38.58 & 61.42 & 50.00 & 54.44 & 65.85 & 61.81 \\
    \quad w/ VCD & \underline{15.87} (\textcolor{red}{+1.58}) & 59.16 (\textcolor{red}{+1.04}) & 48.43 (\textcolor{red}{+1.19}) & 40.16 (\textcolor{red}{+1.58}) & \underline{61.42} (+0.00) & 50.79 (\textcolor{red}{+0.79}) & \textbf{61.11} (\textcolor{red}{+6.67}) & \underline{64.02} (-1.83) & \underline{62.99} (\textcolor{red}{+1.18}) \\
    \quad w/ DoLA & \textbf{19.05} (\textcolor{red}{+4.76}) & 56.54 (-1.58) & 47.24 (+0.00) & \underline{47.24} (\textcolor{red}{+8.66}) & 59.84 (-1.58) & \underline{53.54} (\textcolor{red}{+3.54}) & 57.78 (\textcolor{red}{+3.34}) & 62.80 (-3.05) & 61.02 (-0.79) \\
    \quad w/ OPERA & 14.29 (+0.00) & \underline{61.78} (\textcolor{red}{+3.66}) & \underline{50.00} (\textcolor{red}{+2.76}) & 36.22 (-2.36) & \textbf{65.35} (\textcolor{red}{+3.93}) & 50.79 (\textcolor{red}{+0.79}) & \underline{60.00} (\textcolor{red}{+5.56}) & \underline{64.02} (-1.83) & 62.06 (\textcolor{red}{+0.25}) \\
    \quad w/ ARCD & 14.29 (+0.00) & \textbf{62.83} (\textcolor{red}{+4.71}) & \textbf{50.79} (\textcolor{red}{+3.55}) & \textbf{48.82} (\textcolor{red}{+10.24}) & \underline{61.42} (+0.00) & \textbf{55.11} (\textcolor{red}{+5.11}) & \textbf{61.11} (\textcolor{red}{+6.67}) & \textbf{68.29} (\textcolor{red}{+2.44}) & \textbf{65.75} (\textcolor{red}{+3.94}) \\
    \midrule
    \multicolumn{10}{l}{\textit{Medical Visual-Language Model (Fine-Tuning)}} \\
    \multicolumn{1}{l}{Phi3.5V-Med} & 28.57 & 82.72 & 69.29 & 76.38 & 88.19 & 82.28 & 74.44 & 98.17 & 89.76 \\
    \quad w/ VCD & 33.33 (\textcolor{red}{+4.76}) & \underline{81.68} (-1.04) & 69.69 (\textcolor{red}{+0.40}) & 74.02 (-2.36) & \textbf{88.19} (+0.00) & 81.10 (-1.18) & \underline{75.56} (\textcolor{red}{+1.12}) & 97.56 (-0.61) & \underline{89.76} (+0.00) \\
    \quad w/ DoLA & \underline{47.62} (\textcolor{red}{+19.05}) & 80.10 (-2.62) & \underline{72.05} (\textcolor{red}{+2.76}) & \underline{77.17} (\textcolor{red}{+0.79}) & 86.61 (-1.58) & 81.89 (-0.39) & 73.33 (-1.11) & \underline{98.17} (+0.00) & 89.37 (-0.39) \\
    \quad w/ OPERA & \textbf{50.79} (\textcolor{red}{+22.22}) & 78.53 (-4.19) & 71.65 (\textcolor{red}{+2.36}) & \underline{77.17} (\textcolor{red}{+0.79}) & \underline{87.40} (-0.79) & \underline{82.28} (+0.00) & 65.56 (-8.88) & \textbf{99.39} (\textcolor{red}{+1.22}) & 87.40 (-2.36) \\
    \quad w/ ARCD & 42.86 (\textcolor{red}{+14.29}) & \textbf{89.53} (\textcolor{red}{+6.81}) & \textbf{77.95} (\textcolor{red}{+8.66}) & \textbf{77.95} (\textcolor{red}{+1.57}) & \textbf{88.19} (+0.00) & \textbf{83.07} (\textcolor{red}{+0.79}) & \textbf{81.11} (\textcolor{red}{+6.67}) & \underline{98.17} (+0.00) & \textbf{92.13} (\textcolor{red}{+2.37}) \\
    \bottomrule
    \end{tabular}
    }
    \caption{Performance comparison of MedVQA methods on three datasets. The proposed method (\textit{w/ ARCD}) provides attentional guidance via a segmentation mask and is evaluated against strong baselines and decoding strategies. *On the MIMIC dataset, evaluation is restricted to queries with an identifiable bounding box, where the mask is generated by filling the specified region. }
  \label{tab:sota}%
\end{table*}%

\subsection{Implementation Details}

Our experiments were conducted using 24GB NVIDIA 3090 GPUs, leveraging bfloat16 precision for all training processes. For visual instruction fine-tuning, we initialized the model with pre-trained weights from Phi-3.5-Vision. We employed the AdamW optimizer with a cosine learning rate scheduler, setting the learning rate to 2e-4, the batch size to 256, and training for 1 epoch. For downstream task fine-tuning, we adapted the learning rate to 1e-4, reduced the batch size to 64, and trained for 10 epochs. Only the LoRA layers (rank 64) were updated throughout both phases. 

During inference, our proposed Anatomical Region-Guided Contrastive Decoding is employed as a decoding strategy for both our zero-shot and fine-tuned Phi-3.5V-Med baselines. The core of this strategy is to modulate the influence of anatomical region-guided embeddings relative to normal embeddings at three key stages of generation. This is achieved using parameters $\alpha$ (in Eq. \ref{eq:token_weighting}), $\beta$ (in Eq. \ref{eq:attention_weighting}), and $\gamma$ (in Eq. \ref{eq:logits_weighting}), which control the weighting at the token-level, attention-level, and logits-level, respectively. By adjusting these weights, we can precisely steer the model's final output. A detailed ablation study on the effects of these parameters is discussed in the next section. Following the methodology of LLaVA-Med \cite{LLaVAMedTALi2023}, we assessed performance using accuracy for closed-set questions and recall, defined as the ratio of ground-truth tokens present in the generated response, for open-set questions.

\subsection{Baselines}
For our experiments, we use greedy decoding as the standard methodology to assess the VLMs' inherent performance. We then apply and compare various advanced decoding strategies, including our proposed ARCD, to this baseline. Our baseline models are two variants of Phi3.5V. The first, named \textit{Phi3.5V-Med Zero-shot}, is the model fine-tuned on the general medical vision-language dataset, PubMedVision. The second, \textit{Phi3.5V-Med Fine-Tuning}, is the model further fine-tuned specifically on each downstream task dataset. For a comprehensive performance benchmark, we compare our models against several state-of-the-art methods, which are categorized as follows:

\begin{itemize}

\item \textbf{General VLMs:} We select representative models from the general domain, including LLaVA-1.6 \cite{li2023llava}, Qwen-VL \cite{Bai2023QwenVLAV}, Phi-3.5V \cite{Abdin2024Phi3TR}, and GPT-4o \cite{GPT4TROpenAI2023}.

\item \textbf{Medical VLMs:} We include models specifically tailored for the medical domain, such as LLaVA-Med-7B \cite{LLaVAMedTALi2023} and HuatuoV-34B \cite{huatuogptv}.

\item \textbf{Decoding Strategies:} We apply several plug-and-play decoding strategies to our baseline, including VCD \cite{vcdLeng2023MitigatingOH}, DoLa \cite{DBLP:journals/corr/abs-2309-03883_DOLA}, and OPERA \cite{Huang2023OPERAAH}.

\end{itemize}

\section{Results}
\subsection{Comparisons with Baselines}
As detailed in Table~\ref{tab:sota}, our proposed ARCD strategy, which uses anatomical region masks for guidance, consistently demonstrates accuracy improvements across all three MedVQA datasets. In the zero-shot scenario, our approach achieves pronounced accuracy gains on SLAKE (+5.11\%) and OBScan (+3.94\%). The improvement on the more fine-grained MIMIC dataset is more modest (+3.55\%). This suggests that while the anatomical mask provides correct spatial guidance, the zero-shot model's performance is ultimately constrained by its lack of specialized diagnostic knowledge. In contrast, the MIMIC fine-tuned model, which already possesses this knowledge, fully leverages the anatomical guidance for a substantial +8.66\% improvement. Furthermore, our technique consistently outperforms other decoding strategies like VCD and DoLa. We attribute this to the strong anatomical prior introduced by the segmentation mask, which effectively constrains the model's attention to clinically relevant anatomical structures and reduces hallucinations.

\subsection{Ablation Study}
\label{sec:ablation}
In this section, we conduct ablation studies to: 1) Analyze the causes of hallucinations in MedVLMs; 2) Investigate the impact of different guidance information and parameters; 3) Evaluate the influence of alternative visual prompts.

\begin{figure*}[t]
  \centering
  \includegraphics[width=1\textwidth]{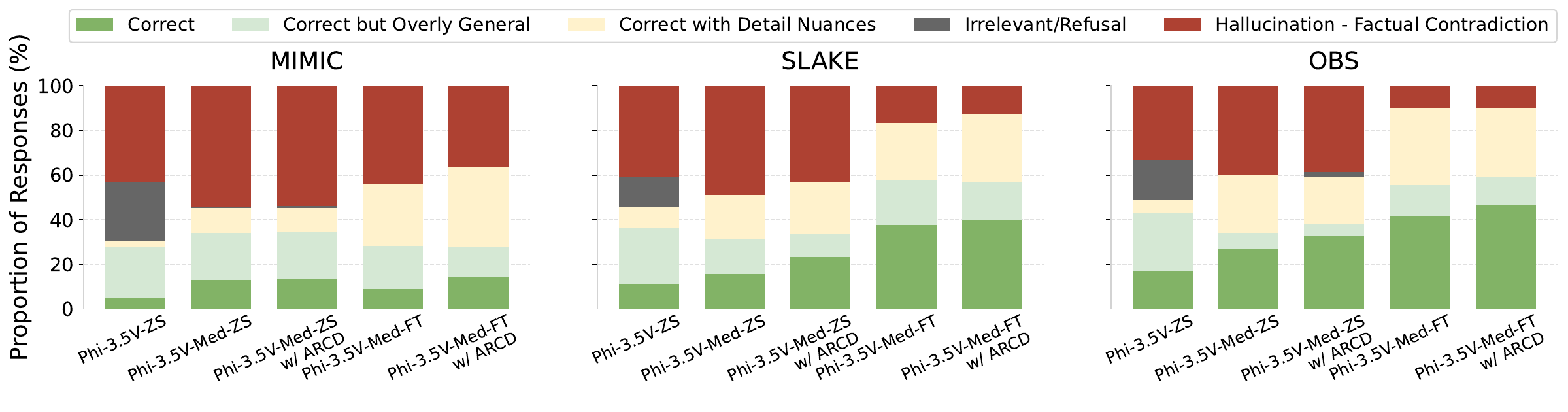}
  \caption{GPT-4o evaluation of model responses under different settings on 250 samples uniformly drawn from three datasets. \textit{Phi-3.5V-ZS} and \textit{Phi-3.5V-Med-ZS} represent the zero-shot results for the base model and the model adapted with PubMedVision. \textit{Phi-3.5V-Med-FT} is the model fine-tuned on the three MedVQA datasets, while \textit{w/ ARCD} denotes our proposed method with attentional masking.}
  \label{fig:param_hallu}
\end{figure*}

\textbf{MedVLM Hallucination Evaluation.} While prior work in the general domain has analyzed and investigated the sources of VLM hallucinations, a unified evaluation standard tailored for the medical domain is lacking. To address this, we qualitatively analyze model outputs, which GPT-4o classifies into five categories: \textit{Correct}, \textit{Correct but Overly General}, \textit{Correct with Detail Nuances}, \textit{Irrelevant/Refusal}, and \textit{Hallucination/Factual Contradiction}, as shown in Figure~\ref{fig:param_hallu}. Our results indicate that fine-tuning on both general (\textit{Phi3.5V-Med-ZS}) and task-specific (\textit{Phi3.5V-Med-FT}) medical data significantly reduces hallucinations and refusals. Notably, our proposed \textit{w/ ARCD} method consistently increases the proportion of fully \textit{Correct} responses over its corresponding baseline in every setting. However, the complexity of the MIMIC dataset, where reports often detail numerous co-occurring diseases, can lead the fine-tuned model to generate more plausible but unverified information. Consequently, such outputs must be subjected to expert review for clinical validation.

\begin{table}[t]
  \centering
  \resizebox{0.9\columnwidth}{!}{
    \begin{tabular}{p{3cm}ccc}
    \toprule
    \multicolumn{1}{l}{Method} & \textbf{MIMIC-Ext-VQA} & \textbf{SLAKE*} & \textbf{OBScan} \\
    \midrule
    \multicolumn{1}{l}{Phi3.5V-Med-ZS} & 47.24 & 50.00 & 61.81 \\
    \quad w/ Label Prompt & 49.60 & 51.97 & 53.14 \\
    \quad w/ Bbox Prompt & \cellcolor[rgb]{ .851,  .851,  .851}\textbf{50.79} & \textbf{55.11} & 64.96 \\
    \rowcolor[rgb]{ .851,  .851,  .851} \quad w/ Mask Prompt & \cellcolor[rgb]{ 1,  1,  1} - & \textbf{55.11} & \textbf{65.74} \\
    \quad w/ Label+Bbox & \textbf{50.79} & 54.33 & 59.44 \\
    \quad w/ Label+Mask & -    & 54.33 & 62.59 \\
    \bottomrule
    \end{tabular}
    }
\caption{Prompt type ablation of Phi3.5V-Med-ZS on medical VQA benchmarks. In the SLAKE*, only a portion of the images have mask annotations. \textbf{Bolded} results are the default setting.}
  \label{tab:info_abl}%
\end{table}%

\textbf{Impact of Guidance Information.} 
While our primary approach uses region mask guidance, other information sources like bounding box or class labels could also influence performance. We thus ablate these components and their combinations, with results in Table \ref{tab:info_abl}. Across all datasets, mask-based guidance provides the most substantial performance boost. Bbox guidance, though less precise, also yields consistent gains, confirming that strengthening the model's focus on specific spatial regions helps mitigate hallucinations. For the MIMIC and SLAKE datasets, all guidance components improved performance. However, on OBScan, using only the class label was detrimental (-8.67\%). We hypothesize this is because our labels are often clinical abbreviations (e.g., ``VD'' as ``Vitreous Hemorrhage''), which may introduce ambiguity or conflict with the visual evidence without precise spatial grounding.

\begin{figure}[t]
  \centering
  \includegraphics[width=1\columnwidth]{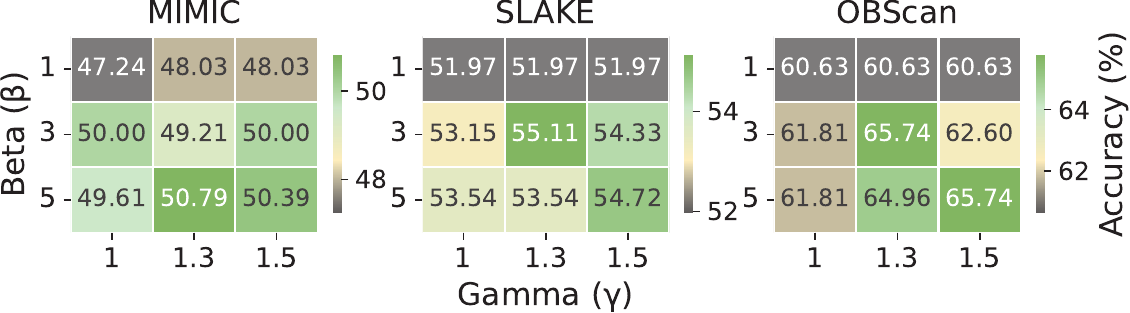}
  \caption{Parameter ablation of $\beta$ and $\gamma$ using the Phi-3.5V-Med zero-shot model, with $\alpha = 0.01$ fixed.}
  \label{fig:param_abl}
\end{figure}

\begin{figure}[t]
  \centering
  \includegraphics[width=1\columnwidth]{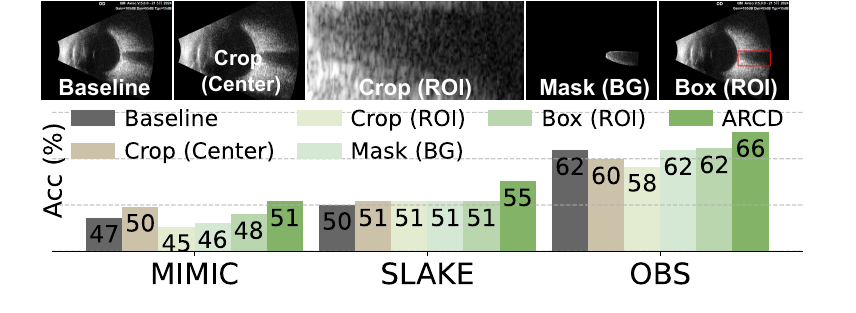}
  \caption{Ablation Study on Visual Prompting: Impact of Visual ROI and Attention on MedVLM Performance.}
  \label{fig:prompt_vp}
\end{figure}

\begin{figure*}[t]
  \centering
  \includegraphics[width=1\textwidth]{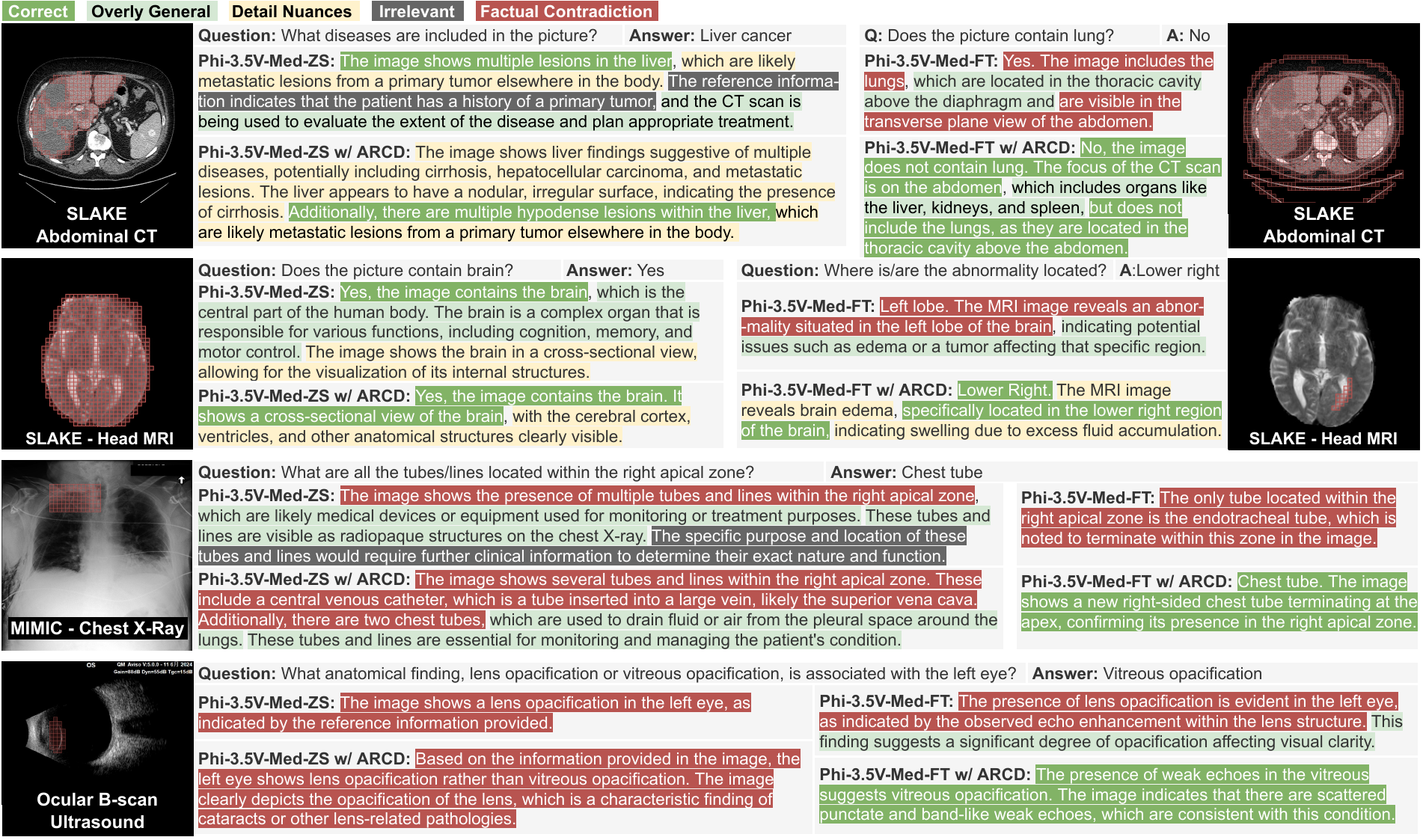}
  \caption{ Qualitative comparison of model-generated responses across four medical imaging modalities: Abdominal CT, Head MRI, Chest X-ray, and Ocular B-scan ultrasound. We compare our proposed method against \textit{Phi-3.5V-Med} zero-shot and fine-tuned settings. Since the global and local masks coincide, only the local mask is visualized as small red boxes.
}
  \label{fig7:case}
\end{figure*}

\textbf{Impact of Control Parameters.} We present a detailed ablation study in Figure~\ref{fig:param_abl} on the control parameters $\alpha$ (in Eq. \ref{eq:token_weighting}), $\beta$ (in Eq. \ref{eq:attention_weighting}), and $\gamma$ (in Eq. \ref{eq:logits_weighting}). Setting $\alpha=1$ disables our ARCD module, meaning the mask $m$ does not affect the input token embedding $c$. Setting $\beta=1$ or $\gamma=1$ disables the corresponding attention-level and logits-level guidance, respectively. For attention-level guidance, performance peaks when $\beta$ is 3 or 5 across datasets, suggesting an optimal $\beta$ balances enhanced focus with preserving broader context. For logits-level guidance, $\gamma=1.3$ is consistently optimal or near-optimal. Since model performance is insensitive to $\alpha$ within the [0, 0.1] range, we defer detailed results to the appendix. These findings validate that our balanced, multi-level guidance is superior to single-level interventions.

\textbf{Impact of Visual Prompting.} To further investigate the effectiveness of our proposed method, we conduct an ablation study on various forms of input visual information, as shown in Figure \ref{fig:prompt_vp}. As the results show, simple cropping or masking methods are clearly outperformed by our carefully designed ARCD, although performance varies slightly across different datasets. On the MIMIC dataset, a simple center crop yields a notable improvement, while other methods show only marginal gains or even a decrease in performance. For the SLAKE dataset, all visual prompting strategies provide a consistent and modest performance boost over the baseline. On the OBS dataset, methods guiding attention to the ROI, particularly using a bounding box, achieve the best results; interestingly, any cropping method harms performance, and we speculate this is because unprocessed textual information, which are lost during cropping, provide a helpful prior. Notably, simply overlaying a bounding box on the relevant region consistently provides a stable performance improvement, as noted in \cite{Shtedritski2023WhatDC}, which indirectly highlights the importance of attention priors for enhancing VLM accuracy and reliability and demonstrates the effectiveness of ARCD.

\subsection{Case Study}
Figure~\ref{fig7:case} showcases a qualitative comparison of the responses generated by our proposed ARCD against those from the Zero-shot Med-Phi3.5V and the Fine-tuned Med-Phi3.5V baselines. The cases are selected from a diverse range of medical imaging modalities, including chest X-ray, abdominal CT, brain MRI, and ocular B-scan ultrasound, with expert-annotated masks provided as a ground-truth reference for the clinically relevant anatomical regions. As observed, the baseline models exhibit distinct failure modes across various tasks. The fine-tuned model, for instance, often contradicts facts by incorrectly identifying an abnormality in the \textit{Left lobe} of the brain instead of the correct \textit{Lower right} and hallucinates the presence of lungs in an abdominal CT scan. Similarly, both baseline models misidentify the condition in an ocular ultrasound as \textit{lens opacification} when it is \textit{vitreous opacification}. In contrast, our ARCD strategy consistently provides the correct answers in these challenging cases. This demonstrates that by leveraging anatomical masks to ground the model's attention, our approach effectively mitigates factual errors and hallucinations, leading to more accurate and reliable diagnostic responses.

\section{Conclusion}
In conclusion, hallucination remains a critical barrier to the clinical adoption of Medical Vision-Language Models (MedVLMs). To address this, we introduced Anatomical Region-Guided Contrastive Decoding (ARCD), a novel, training-free strategy that steers language generation using specified anatomical regions. By applying a three-tiered, contrastive guidance mechanism, our method provides fine-grained control over the model's visual attention, directly tackling the core problem of poor visual grounding. This plug-and-play approach avoids expensive retraining while offering more reliable correction than existing inference-time methods. Extensive experiments across diverse datasets validate its effectiveness and generalization. Our work offers a new approach for mitigating hallucinations in medical vision-language models, driving progress in the field.

\section{Acknowledgments}
This work was supported in part by the National Natural Science Foundation of China under Grants 62192730, 62192734, 62577041, the National Science and Technology Major Project under Grant 2022ZD0117103, the Outstanding Youth Science Foundation of Shaanxi Province under Grant 2025JC-JCQN-083, the Key Research and Development Program of Shaanxi Province under Grants 2025CY-YBXM-047 and 2024GX-YBXM-140, and the CCF-Huawei Populus Grove Fund under Grant CCF-HuaweiFM202507.
\bibliography{aaai2026}

\clearpage

\appendix

\onecolumn 
\begin{center}
  {\Large \bfseries Supplementary Material for \\ Anatomical Region-Guided Contrastive Decoding\par}
  \vspace{1.5em}
\end{center}

\section{Dataset Details and Construction}
\label{sec:appendix_dataset}

\subsection{Dataset Overview}
\label{sec:appendix_dataset_stats}
Table 3 summarizes the key statistics of the three downstream task datasets used in this study.

\begin{table}[htbp]
  \centering
  \label{tab:appendix_dataset_stats}
  \resizebox{0.9\textwidth}{!}{%
    \begin{tabular}{lrrlp{10.465em}}
    \toprule
    \textbf{Dataset} & \multicolumn{1}{l}{\textbf{\#Images}} & \multicolumn{1}{l}{\textbf{\#QA-Pairs (Train / Val / Test)}} & \textbf{Type} & \multicolumn{1}{l}{\textbf{Usage}} \\
    \midrule
    SLAKE & 642   & 5,972 / 1,053 /1,061 & Radiology & Fine-tuning \& Evaluation \\
    OBScan & 273   & 1,695 / 210 / 254 & Radiology & Fine-tuning \& Evaluation \\
    MIMIC & 53,356 & 44,686 / 44,686 / 2,134 & Radiology & Fine-tuning \& Evaluation \\
    PubMedVision & 1,009,700 & 647,031 & Comprehensive & Fine-tuning \\
    \bottomrule
    \end{tabular}%
  }
  
  \caption{\textbf{Dataset statistics}, including the number of images, QA-pairs, and data types are detailed. }
\end{table}%

\subsection{OBScan Dataset Construction}
\label{sec:appendix_obscan_construction}
OBScan is an ocular B-ultrasound visual question answering dataset we constructed for this research.
\paragraph{Data Format} Each sample consists of an image and a corresponding JSON file. The structure of the JSON file, which contains the QA pairs and anatomic region and bounding box, is shown in Listing \ref{lst:data_format}.

\begin{figure*}[!h]
\begin{minipage}{\textwidth}
\begin{lstlisting}[frame=single]
{
    "split": "test",
    "question": "What condition is indicated by the presence of increased echogenicity in the lens of the left eye?",
    "question_type": "attribute",
        "template_arguments": {
            "object": {"0": "lens"},
            "category": {"0": "cataract"},
            "attribute": "increased echogenicity",
            "bbox": {
                "0": [
                    193.39,
                    212.36,
                    250.86,
                    366.95
                ]
            },
            "viewpos": "left eye",
            "mask_value": {"0": 128}
        },
    "answer": ["cataract"],
    "explanation": "The increased echogenicity observed in the lens suggests the presence of a cataract. This condition is characterized by clouding of the lens, which can lead to visual impairment.",
    "answer_type": "closed",
    "image_ids": ["Img_5638_OS"],
    "subject_id": "Z0478897"
}
\end{lstlisting}
\end{minipage}
\caption{The data format for each sample in the OBScan dataset.}
\label{lst:data_format}
\end{figure*}

\paragraph{Prompt for QA Pair Generation} Given that the annotations in MIMIC-Ext-VQA and SLAKE are limited to short answers insufficient for in-depth analysis, we employed GPT-4o to automatically generate detailed explanations from expert-written reports. The core prompt is shown in Listing \ref{lst:qa_gen_prompt}.

\begin{figure*}[!h]
\begin{minipage}{\textwidth}
\begin{lstlisting}[frame=single]
You are a concise medical explanation assistant. Your task is to generate a brief, accurate 
explanation for a medical image question-answer pair, incorporating the provided current 
answer. The explanation must be in English, start with the answer (capitalized first letter), 
followed by a simple explanation, and be no more than 64 words.

# Example
[Input]
All QA pairs for the image: What modality is used to take this image? ct. Which part of the 
body does this image belong to? chest. What is the main organ in the image? spinal cord.
Current Question: What is the main organ in the image?
Current Answer: spinal cord
Explanation for Current Answer: 

[Output]
Spinal cord. The image shows a cross-section of the chest, and within the vertebral column, the spinal cord is clearly visible.

# User Query Template
[Input]
All QA pairs for the image: {all_qa_pairs_str}
Current Question: {question}
Current Answer: {answer}
Explanation for Current Answer: 

\end{lstlisting}
\end{minipage}
\caption{The prompt for generating QA pairs for the SLAKE dataset. For the SLAKE dataset, which lacks expert-written reports, we created a textual context for each image by concatenating all its corresponding QA pairs. We excluded pairs where the answer was ``No," as they provide no informative content. For the MIMIC dataset, we used the provided reports directly.}
\label{lst:qa_gen_prompt}
\end{figure*}

\paragraph{Annotation and Data Privacy} The segmentation masks for anatomical structures and pathological regions in the OBScan dataset were cross-annotated and verified by two attending ophthalmologists, each with over five years of clinical experience, to ensure annotation accuracy and professionalism. The data, including all images, reports, and QA pairs, was \textit{shared under a collaborative research agreement after being fully de-identified to remove personal information such as names, medical records, and contact details}.

\subsection{Note on MIMIC-Ext-VQA}
\label{sec:appendix_mimic_subset}
Our evaluation on MIMIC-Ext-VQA is based on a subset of queries with identifiable bounding boxes. This bounding box information originates from the Chest-imagenome \cite{Wu2021ChestID_ChestImaGenome} dataset, which itself provides corresponding bounding box annotations for a portion of questions related to specific lesion locations. We utilized these existing bounding boxes to generate our attention guidance masks.

\clearpage

\section{Implementation Details}
\label{sec:appendix_exp_details}

\subsection{Hyperparameter Settings}

\subsubsection{Ablation Study for Contrastive Decoding Parameters ($\alpha, \beta, \gamma$)}
\label{sec:appendix_ablation_hyperparams}
Figure \ref{fig:param_abl} in the main text illustrates the impact of the key parameters $\beta$ and $\gamma$ on the zero-shot model's performance. We conducted further ablation studies for the fine-tuned model, with the results presented in Table 4. The experiments indicate that $\alpha$ has a limited effect on performance; consequently, we set a fixed value of $\alpha=0.01$. Based on these ablation studies, we report results in Table \ref{tab:sota} that are robust and stable, although not necessarily the absolute best-performing configuration.

\begin{table}[htbp]
  \centering
  \label{tab:hyperparameter_ablation}%
  \resizebox{\textwidth}{!}{%
    \begin{tabular}{l|rrrrr|rrrrr|rrrrr}
    \toprule
    \textbf{\#Param.} & \multicolumn{5}{c|}{\textbf{OBScan}}  & \multicolumn{5}{c|}{\textbf{SLAKE}}   & \multicolumn{5}{c}{\textbf{MIMIC}} \\
    \midrule
    $\alpha$ in Eq(\ref{eq:token_weighting}) & 0     & 0.1   & 0.01  & 0.01  & 0.01  & 0     & 0.1   & 0.01  & 0.01  & 0.01  & 0     & 0.1   & 0.01  & 0.01  & 0.01 \\
    $\beta$ in Eq(\ref{eq:attention_weighting})  & 3     & 3     & 3     & 5     & 5     & 3     & 3     & 3     & 5     & 5     & 3     & 3     & 3     & 5     & 5 \\
    $\gamma$ in Eq(\ref{eq:logits_weighting})& 1.3   & 1.3   & 1.3   & 1.3   & 1.5   & 1.3   & 1.3   & 1.3   & 1.3   & 1.5   & 1.3   & 1.3   & 1.3   & 1.3   & 1.5 \\
    \midrule
    Accuracy & 0.9055 & 0.9055 & 0.9055 & 0.9173 & \textbf{\textit{0.9213}} & 0.8228 & 0.8228 & 0.8228 & \textit{\textbf{0.8307}} & \textit{\textbf{0.8307}} & \textit{0.7795} & \textit{0.7795} & \textit{0.7795} & \textbf{0.7913} & \textbf{0.7913} \\
    \bottomrule
    \end{tabular}%
  }
    \caption{Ablation study for the hyperparameters $\alpha$, $\beta$, and $\gamma$ on the OBScan, SLAKE, and MIMIC datasets. The table reports the final accuracy for each parameter configuration. Numbers in \textbf{bold} and \textit{italics} represent the best performance and the results reported in Table 1, respectively.
}
\end{table}%

\subsubsection{Hyperparameters for Baseline Decoding Strategies}
\label{sec:appendix_baseline_hyperparams}
We adopted the default hyperparameter settings recommended in the original papers for all baseline decoding strategies (VCD, DoLa, OPERA). These parameters are detailed in Table \ref{tab:appendix_baseline_params}. 

\begin{table}[htbp]
  \centering

  \resizebox{\textwidth}{!}{%
    \begin{tabular}{c|lllp{38.465em}}
    \toprule
    Type  & Method & Param. & Value & Description \\
    \midrule
    \multirow{3}[2]{*}{VLMs} & HuatuoV & \multirow{3}[2]{*}{max\_token} & \multicolumn{1}{r}{1024} & \multirow{3}[2]{*}{Default Setting.} \\
          & QwenVL &       & \multicolumn{1}{r}{512} & \multicolumn{1}{l}{} \\
          & LlaVA-Med &       & \multicolumn{1}{r}{1024} & \multicolumn{1}{l}{} \\
    \midrule
    \multicolumn{1}{c|}{\multirow{11}[8]{*}{\makecell[c]{Decoding\\Strategies}}} & \multicolumn{4}{l}{\textit{\# All decoding strategies use max\_token=256 and num\_beam=1, except for OPERA.}} \\
\cmidrule{2-5}          & \multirow{4}[2]{*}{VCD} & $\alpha$ & \multicolumn{1}{r}{0.35} & Contrastive strength coefficient. Controls the output distribution weight between original and noisy images. Higher value = stronger contrast. \\
          &       & $\beta$  & \multicolumn{1}{r}{0.03} & Adaptive plausibility threshold. Controls the truncation of the next-token distribution. Larger $\beta$ indicates more aggressive truncation. \\
          &       & $\gamma$ & \multicolumn{1}{r}{0.1} & Diffusion noise strength. Controls the amount of noise added in each step. \\
          &       & T & \multicolumn{1}{r}{250} & Diffusion noise steps. The total number of noise-adding steps. \\
\cmidrule{2-5}          & \multirow{5}[2]{*}{OPERA} & $\sigma$ & \multicolumn{1}{r}{40} & Configurable scaling factor. \\ 
          &       & Ncan  & \multicolumn{1}{r}{4} & Number of attention candidates. \\
          &       & $\alpha$ & \multicolumn{1}{r}{0.85} & Penalty weight. Weights the Over-Trust Logit Penalty that is subtracted from the model logits. \\
          &       & r     & \multicolumn{1}{r}{15} & Retrospection threshold. \\
          &       & num\_beam & \multicolumn{1}{r}{3} & \multicolumn{1}{r}{} \\
\cmidrule{2-5}          & DoLA  & dola\_layers & 2, 4, 6, 8 & Specifies the Transformer layers used for contrast during DoLA decoding. \\
    \bottomrule
    \end{tabular}%
  }
    \caption{Hyperparameter settings for Vision-Language Models (VLMs) and various decoding strategies.}
  \label{tab:appendix_baseline_params}%
\end{table}%

\subsection{Implementation Details for Visual Prompting}
\label{sec:appendix_visual_prompt_details}
In Figure 5 of the main text, we investigated the impact of different visual prompting strategies. The specific implementation details for each strategy are as follows:
\begin{itemize}
    \item \textbf{Center Crop:} The central 80\% of the image area was cropped and then resized back to the original dimensions.
    \item \textbf{ROI Crop:} Only the ROI area was cropped. The cropped sub-image was then resized to the original dimensions.
    \item \textbf{Mask Background:} Cover the area outside of the expert annotations with black.
    \item \textbf{Overlay BBox:} A red, 2-pixel wide, semi-transparent (100\% alpha) bounding box was overlaid on the region of interest (ROI) in the image.
    
\end{itemize}

\subsection{Evaluation Details}
\label{sec:appendix_eval_details}

\subsubsection{Prompt for GPT-4o Evaluation}
\label{sec:appendix_gpt4_prompt}
To enhance the transparency and reproducibility of our evaluation method (as shown in Figure \ref{fig:param_hallu}), the exact prompt used to instruct GPT-4o for response classification is provided in Listing \ref{lst:gpt4o_prompt}.

\begin{figure*}[!h]
\begin{minipage}{\textwidth}
\begin{lstlisting}[frame=single]
You are an expert AI evaluator. Classify the "Model Prediction" into ONE of the following five categories based on the provided ground truth.

[Evaluation Flow]
1. First, check if the "Model Prediction" contains the key information from the "Ground Truth Short Answer".
2. Based on that, and by comparing the prediction's details to the "Ground Truth Explanation", select a category from the list below.

[Categories]
1.  **Correct**: The prediction contains the short answer AND its details are perfectly aligned with the ground truth explanation.
2.  **Correct but Overly General**: The prediction contains the short answer, BUT is vaguer or lacks specific details found in the ground truth explanation.
3.  **Correct but Incorrectly Specific**: The prediction contains the short answer, BUT incorrectly adds or modifies details (e.g., wrong size, location, number).
4.  **Hallucination - Factual Contradiction**: The prediction does NOT contain the short answer AND directly contradicts the ground truth.
5.  **Irrelevant/Refusal**: The prediction is off-topic or a refusal to answer.

[Input]
- Ground Truth Short Answer: {gt_short_answer}
- Ground Truth Explanation: {gt_explanation}
- Model Prediction: {model_prediction}

[Instruction]
Respond with ONLY the category label chosen from the list above.
\end{lstlisting}
\end{minipage}
\caption{The prompt used for GPT-4o evaluation.}
\label{lst:gpt4o_prompt}
\end{figure*}


\section{Additional Experiments and Analysis}
\label{sec:appendix_additional_exp}

\subsection{Annotation Strategy: Expert Masks vs. Automated Masks}
\label{sec:appendix_mask_strategy}
To validate the ``plug-and-play" nature of our ARCD method and to evaluate its effectiveness in the absence of expert annotations, we conducted a comparative experiment. For the MIMIC dataset, which has clear expert organ annotations and available public models, we utilized the PSPNet segmentation model from the torchxrayvision project \cite{Cohen2021TorchXRayVisionAL} to generate organ masks, taking the maximum enclosing rectangle as the bounding box. The prediction accuracy of these bounding boxes is shown on the left side of the table below. The results on the right show the performance in both zero-shot and fine-tuning settings when these auto-generated bounding boxes replace the original expert annotations. The low precision at IoU $> 0.75$ is an expected result of a granularity mismatch: the model predicts broad categories like \textit{right lung}, whereas the ground truth annotations are more specific, such as \textit{right mid lung zone}. Although the regions overlap, the resulting IoU is low. Nevertheless, this does not hinder the overall performance gains.

\begin{table}[htbp]
  \centering
  \resizebox{0.8\textwidth}{!}{%
    \begin{tabular}{llrrr|lrr}
    \toprule
    \textbf{Method} & \multicolumn{4}{c|}{\textbf{MIMIC for Bbox}} & \multicolumn{3}{c}{\textbf{MIMIC for VQA}} \\
    \cmidrule(r){2-5} \cmidrule(l){6-8}
          & Metric & \multicolumn{1}{l}{IoU=0.25} & \multicolumn{1}{l}{IoU=0.5} & \multicolumn{1}{l|}{IoU=0.75} & Metric(\%) & \multicolumn{1}{l}{Zero-Shot} & \multicolumn{1}{l}{Fine-Tuning} \\
    \midrule
    \multicolumn{1}{l}{\multirow{3}[2]{*}{\textbf{PSPNet}}} & Precision & 0.7035 & 0.3761 & 0.0575 & Open  & 15.87 & 57.14 \\
          & Recall & 0.8503 & 0.7522 & 0.3171 & Closed & 58.64 & 83.25 \\
          & F1    & 0.7700 & 0.5015 & 0.0974 & Overall & 48.03 & 76.77 \\
    \bottomrule
    \end{tabular}%
  }
  \caption{\textbf{Performance evaluation of automated organ detection and the VQA performance using detected bounding boxes to replace expert annotations.} We use detection metrics for evaluation as MIMIC lacks segmentation ground truth. The left side details bounding box detection performance (Precision, Recall, F1-score) at varying IoU thresholds. The right side shows Visual Question Answering (VQA) accuracy for different question types under zero-shot and fine-tuning settings.}
  \label{tab:pspnet_performance}%
\end{table}%


\end{document}